# e-Inu: Simulating A Quadruped Robot With Emotional Sentience

[1]Abhiruph Chakravarty, [2]Jatin Karthik Tripathy, [1]Sibi Chakkaravarthy S, [3*]Aswani Kumar Cherukuri
[4]S. Anitha  [5]Firuz Kamalov  [6]Annapurna Jonnalagadda
[1]School of Computer Science and Engineering and Center of Excellence, Artificial Intelligence and Robotics (AIR)
[2]Cognitive Systems, University of Potsdam, Germany
[4]School of Electronics Engineering
VIT-AP University, Andhra Pradesh, India
[3]School of Information Technology, Vellore Institute of Technology, Vellore, India
[5]Dept. of Electrical Engineering, Canadian University Dubai, UAE
[6]School of Computer Science and Engineering, Vellore Institute of Technology, Vellore, India.
(*Corresponding author: cherukuri@acm.org)

**Abstract**

Quadruped robots are currently used in industrial robotics as mechanical aid to automate several routine tasks. However, presently, the usage of such a robot in a domestic setting is still very much a part of the research. This paper discusses the understanding and virtual simulation of such a robot capable of detecting and understanding human emotions, generating its gait, and responding via sounds and expression on a screen. To this end, we use a combination of reinforcement learning and software engineering concepts to simulate a quadruped robot that can understand emotions, navigate through various terrains and detect sound sources, and respond to emotions using audio-visual feedback. This paper aims to establish the framework of simulating a quadruped robot that is emotionally intelligent and can primarily respond to audio-visual stimuli using motor or audio response. The emotion detection from the speech was not as performant as ERANNs or Zeta Policy learning, still managing an accuracy of 63.5%. The video emotion detection system produced results that are almost at par with the state of the art, with an accuracy of 99.66%. Due to its "on-policy" learning process, the PPO algorithm was extremely rapid to learn, allowing the simulated dog to demonstrate a remarkably seamless gait across the different cadences and variations. This enabled the quadruped robot to respond to generated stimuli, allowing us to conclude that it functions as predicted and satisfies the aim of this work.

Keywords: Automated Gait Generation, Deep Learning, Emotion Recognition, Reinforcement Learning, Robot Simulation.

## 1. Introduction

Development in robotics has been a fundamental stepping-stone in the progression of humanity. However, unlike most new technologies, robotics is still mostly limited to industrial or professional use, with the rather minuscule exceptions of home automation, smart home cleaners, etc. With that in mind, the need for pets, more specifically dogs, has grown profusely – both for emotional and safety needs. It can be quite difficult to care for an organic pet in a modern fast-paced life. To that end, we focus our aim on simulating an artificially intelligent quadruped robot [37] using PyBulet, modelled to serve the domestic needs of a sentient, albeit artificial, dog. Three fundamental problems accompany the development of the said robot, viz., emotion analysis, environmental awareness, and automated gait generation. The objectives are

1. To design a system that automates the detection of underlying emotions in speech, tonality, and facial expressions of humans present in the scene,
2. To detect the direction of sound sources and obstacles and,
3. To generate the necessary gait to travel towards the sound source
4. To generate the required audio-visual responses for the situation.

This research discusses the existing ideas or models in the fields of Convoluted Neural Networks (CNN), Recurrent Neural Networks (RNN), Reinforcement Learning (RL), and some software engineering to meet the end goals, such as the use of TDoA. The primary goal is to focus on the emotional aspect of having a pet dog, not the guardian aspect for this paper. A point to note is that this paper does not develop the robotics of the state-of-the-art machines already established in the market. This paper discusses an alternate pathway in which these quadruped machines can be used in a domestic setting, developing the more utopian Neuromancer [1] version of the future. To this end, we endeavour to make a quadruped system capable of a certain degree of emotional intelligence, allowing the robot to feel much more natural.

## 2. Review of Previous Related Work

A major inspiration in the research has been similar quadruped robots with more industrial use cases such as the ANYmal [2] by ANYbotics, Spot [3] by Boston Dynamics, and AlienGo [4] by Unitree Robotics. These quadrupeds were designed to be mainly robust and cover different scenarios that might crop up in an industrial scenario. However, e-Inu implements several key aspects common to all quadrupeds: gait generation and optimization, obstacle avoidance, and route planning. A similar project on the ground of quadruped bots was carried out by Marc Raibert, Blankespoor, Nelson et al.[5], Big Dog, whose goal was to make autonomous quadruped robots that resembled dogs.

One of the recent papers that we explored was by Deng et al.[6] Sparse Autoencoder-based Feature Transfer Learning for Speech Emotion Recognition. According to their findings, training and test data utilized for system development in speech emotion recognition typically fit each other precisely, but additional 'similar' data may be accessible. Transfer learning enables the use of such similar data for training despite underlying differences to improve the performance of a recognizer. Their research provided a sparse autoencoder technique for feature transfer learning for speech emotion identification, which learns a common emotion-specific mapping rule from a limited margin of labelled data in a target domain. Then, newly reconstructed data were obtained using this method on emotion-specific data from a different domain. The experimental findings on six typical databases demonstrated that their approach greatly outperforms learning each source domain independently. The basic idea behind the sparse autoencoder-based feature transfer learning method was to use a single-layer autoencoder to find a common structure in small target data. Then apply that structure to reconstruct source data to complete useful knowledge transfer from source data into a target task. They used the reconstructed data to create a speech emotion identification engine for a real-world problem presented by the Interspeech 2009 Emotion Challenge. The proposed technique effectively transfers knowledge and improves classification accuracy, according to experimental results with six publicly accessible corpora.

In the work of Lu et al. [7], as a downstream task, they proposed using pre-trained features from end-to-end ASR models to perform speech sentiment analysis. End-to-end ASR features that use both acoustic and text information from speech produced encouraging results. As the sentiment classifier, an RNN with self-attention was used, which also gave an accessible visualization using attention weights to help comprehend model predictions. The IEMO-CAP dataset and a new large-scale speech sentiment dataset SWBD-sentiment were employed for evaluation. With over 49,500 utterances, they increased the state-of-the-art accuracy on IEMO-CAP from 66.6% to 71.7% and reached an accuracy of 70.10% on SWBD-sentiment. As a result, it was proved that pre-trained features from the end-to-end ASR model are useful for sentiment analysis. However, the work of Lu et al. [7] was the fundamental motivation for our model to analyze emotions from audio/ speech. Their work demonstrated a spoken emotion identification system based on a recurrent neural network (RNN) model taught by a fast learning algorithm. It considered the long-term context influence as well as the unpredictability of emotional label expressions. A robust learning method using a bidirectional long short-term memory (BiLSTM) model was used to extract a high-level representation of emotional states in terms of their temporal dynamics.

To avoid the ambiguity of emotional labels, it was assumed that the label of each frame was viewed as a sequence of random variables so that all frames in the same speech were mapped into the same emotional label. The proposed learning algorithm was then used to train the sequences. When compared to the DNN-ELM-based emotion identification system utilized as a baseline, the suggested emotion recognition system improved its weighted accuracy by up to 12%. Their method revealed how recurrent neural networks and maximum-likelihood-based learning techniques might be used to improve emotion recognition. This is also when we learned about Mel Frequency Cepstral Coefficients and their application as a feature extractor from audio. We discovered the works of Mermelstein [8], Davis and Mermelstein [9], and Bridle and Brown [10] as we delved deeper.

Maghilnan and Kumar [11] talked about a sentiment analysis algorithm that uses features taken from the voice stream to discern the moods of the speakers in the conversation. Their process included pre-processing with VAD, implementing their Speech Recognition System, implementing their Speaker Recognition System with MFCC, and ultimately implementing their Sentiment Analysis System. Their research offered a generalized model that takes an audio input including a discussion between two persons and examines the content and identity of the speakers by automatically translating the audio to text and performing speaker recognition. The system performed well with the artificially generated dataset; they were working on gathering a larger dataset and boosting the system's scalability. Though the system was accurate in understanding the sentiment of the speakers in a conversational discussion, it could only manage conversations between two speakers speaking one at a time.

Moving on to the next part of the emotion recognition system, facial emotion detection, several works have explored this particular problem. One of the first works in detecting emotion from images was done by Gajarla and Gupta [12]. The approach used in this methodology relied upon the large pre-trained object detection model VGG16 [13] which allowed for the facial detection model to be quickly trained since using VGG16 meant that the model already knew how to pick up cues from an image. This transfer learning method allowed Gajarla and Gupta [12] to work very well by replacing the final layer of the VGG16 model for a different classification layer. The image features were once extracted using VGG16; we passed to an SVM that allowed the overall model to be trained on the task of detecting emotions. This work also explored different architectures such as using the One vs All SVM approach with different VGG16 weights as well as using ResNet50. Gajarla and Gupta [12] obtained impressive results of 73% using the resNet50 model on a dataset consisting of images scraped from "Flickr". One of the main issues faced during this work was the fact that since the images were scrapped from Flickr, the labelling of the images is quite subjective and the ambience and lighting of an image could very easily throw off the model.

Another approach to facial emotion detection was carried out by Dachapally [14] using representational autoencoders units (RAU). The autoencoders allow for a unique representation to be formed for different emotions allowing the images to be easily differentiated. According to their findings, the use of autoencoders to form generalized encoded features of the different emotions since the model looks at different faces whilst training. This methodology also worked well since Dachapally [14] used the JAFFE dataset which has 215 images of 10 different female models posing for 7 emotions. All the images in the training set were of Japanese women, so, all the samples come from the same ethnicity and of the same gender. This fact allows the autoencoders to perform daily decently compared to the transfer learning methodologies discussed by Gajarla and Gupta [12].

The work done by Bargal et al.[15] used an approach very different from the aforementioned approach to the facial emotion recognition problem. This approach used a system that passed the image through three different pre-trained image models, namely VGG13, VGG16, and ResNet50. This parallelization during the extraction of the features from the image model allows the images to be broken down to different levels, resulting in much richer features when compared to using just one pre-trained model. The features from these three models are then individually passed through Signed Square Root (SSR) and L2 Norm before being combined again. This work used the EmotiW'16 Dataset and additional data collected by crawling the web for images tagged with emotion classes. Using this methodology of parallelizing the image feature extraction, Bargal et al. [15] achieved 56.66% on the test dataset, improving over the baseline by a little over 16%. This work also showcased the differences in using a single pre-trained model over parallelization during feature extraction, with the parallel model achieving around 2% more accuracy when using just the VGG16 model.

Several methods have been implemented to optimize both the smoothness and the speed of a gait of a quadruped robot. The initial approach for this goal was heavily dependent on human observations to plan out the path of the leg that offered the best leg movement. Recently, however, the approaches have started implementing deep reinforcement learning to allow for the model to be created without much human intervention.

One of the first approaches to gait analysis, Hengst et al. [16] was dependent on human observations to make a fixed path for the leg to move. This fixed path was configured using three aspects of the gait - movement, speed, and leg stance. These parameters allowed [16] to form the four corners of a rectangle, Figure 1, which laid out the gait's locus and the ability to align the rectangular locus with the robot's body. The robot's leg was moved about the locus using a series of waypoints that were generated per the movement parameters. These waypoints were divided into two equal groups - one for the groundstroke and the other for the rest of the rectangle.

Shortly after, Kim and Uther [17] improved on the Rectangular Locus method by creating a new quadrilateral walk locus described by the four offsets from the original locus, Figure 2, and the speed of the robot's feet around the new locus is the same throughout. This new gait was inherently smoother due to the increase in the number of locus points, the optimization of the locus was to employ Powell's (direction set) method for multidimensional minimization, as outlined in Press et al. [18].

The previous methods can produce satisfactory results, but require a lot of time and human resources for the path to be properly understood and configured to allow for smooth movement of the leg and the robot overall. Clune, Beckmann, Ofria, et al. [19] came up with a new generative encoding methodology for evolving neural networks - HyperNEAT. HyperNEAT was especially impressive compared to the previous methods since it required absolutely no manual tuning since the neural networks eventually found a possible solution to the gait. An added advantage was that generative encoding can more easily reuse phenotypic modules resulting in better leg coordination.

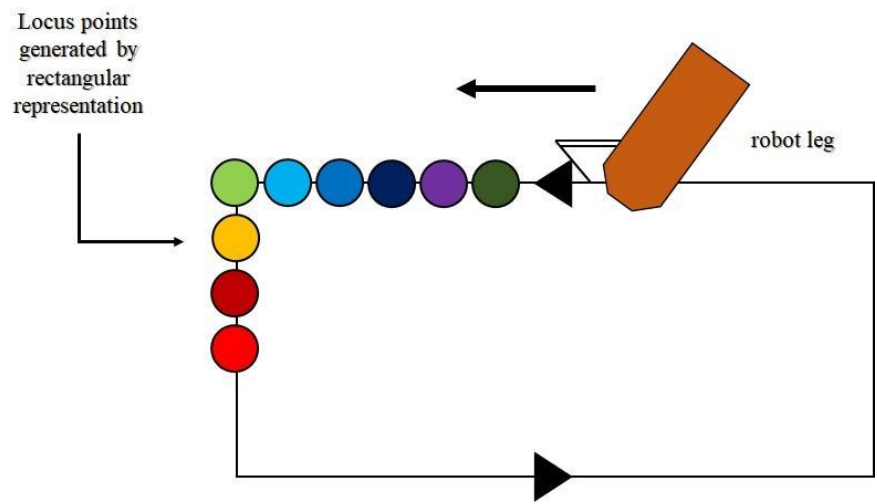

Figure 1. Rectangular Locus (adapted from [16])

Owaki & Ishiguro [20] followed an entirely different ideology compared to the previous methods. Their approach was based on the fact that the movement of the limbs on a quadruped follows a repeating pattern that can be simulated using a "Central Pattern Generator" (CPG) model. Past experiments involving decerebrate cats indicate that cats also have a biological CPG in their spinal cord [21], [22]. This method also allowed for spontaneous gait transition, from walking to trotting to cantering to galloping, which could be achieved by changing one parameter related to speed while maintaining balance and having a low movement cost.

Lodi, Shilnikov, and Storace [23] also worked on the fact that quadrupeds follow a CPG for interlimb communication and synchronization. They proposed a method for designing and analyzing

CPGs, based on multi-parameter bifurcation theory using a recently proposed software tool called CEPAGE [24]. The method is applied to two CPGs, one bio-inspired and one purely synthetic. In both cases, the analysis of the CPGs allows for a way to easily obtain different gait sequences by tweaking the bifurcation parameters. While the proposed method involved some level of human interference as it is mainly a rule-based designing approach, CPGs produce rhythmic patterns allowing for much easier analysis of the gait and the possibility of spontaneous gait transitions.

This paper discusses the development of a domestic-savvy one. The robot in discussion can understand emotions from visual and auditory stimuli and respond to the same with various motor and auditory responses, along with an LCD to make humans more receptive to it.

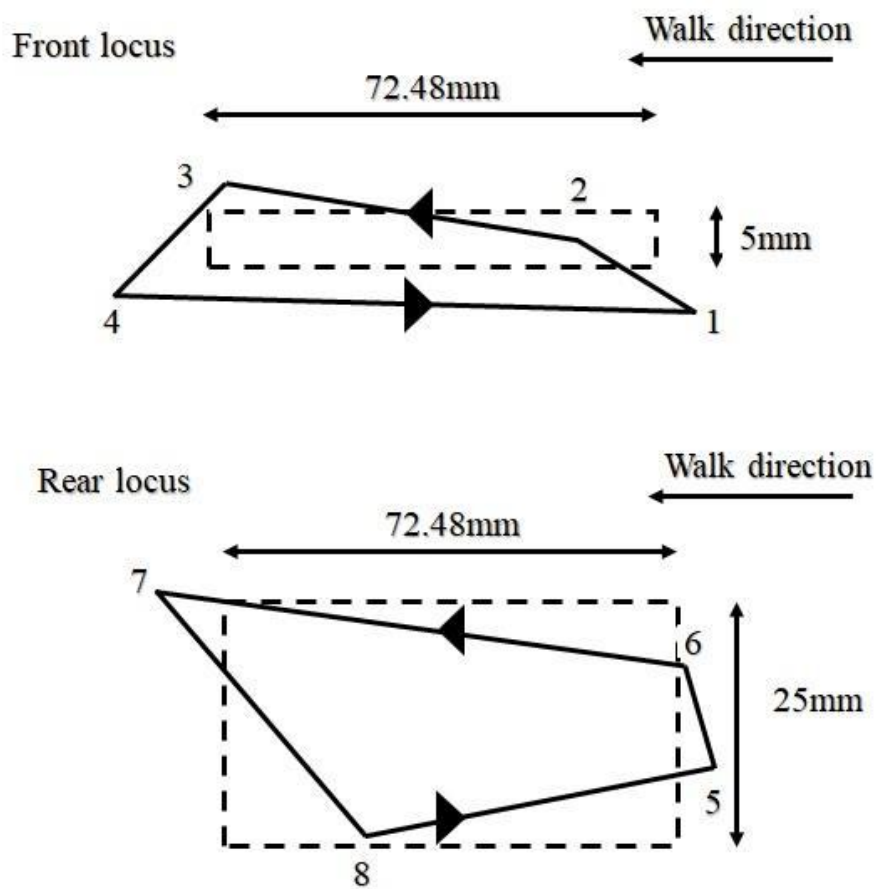

Figure 2. Optimized Rectangular Locus (adapted from [17])

## 3. The proposed Architecture: e-Inu

The e-Inu architecture incorporates four discrete modules to achieve the desired goals. The Emotion detection module is responsible for all aspects of the detection of emotion both audio and visual. The TDoA module gives the robot dog positional awareness. The gait generation module is responsible for generating the gait and mapping the robot's movement from one position to another.

And finally, the audio-visual feedback module helps the robot communicate back with the human and the environment. The components are discussed in the following segments.

## 3.1 Emotion Detection

The emotion detection module considers both the audio and visual inputs as external stimuli and uses separate deep learning networks to compute the same. For the emotion analysis on audio features, we use Mel Frequency Cepstral Coefficients that are extracted from the generated audio and pass it through a pair of LSTM layers sandwiched by four dense layers, two on each side. As for the emotion analysis on video features, we use Haar Cascade to detect facial structures, OpenCV, and a no-top version of VGG16 trained on places365, whose inputs are flattened and passed through dense layers.

MFCCs jocularly referred to in the academic circle as the "Most-Frequently Considered Coefficients", are usually the one-does-it-all when it comes to audio data processing. Any sound produced by humans is dictated by the form of their vocal tract, as also the tongue, teeth, etc. If this form is precisely identified, every sound generated by humans can be appropriately described. The envelope of the temporal power spectrum of a speech signal depicts the vocal tract, and MFCC represents just that. The first thirteen coefficients, or the lower dimensions of MFCC, are considered features representing the aforementioned spectral envelope. The higher dimensions express further details about the spectral features. For various phonemes, envelopes are sufficient to express the difference, allowing us to recognize phonemes using MFCC. But in our case, given that more data is good data, we experimentally arrived at the inference that the forty-dimensional features of MFCCs do the best job.

MFCC is a time series, as data is represented sequentially with the y-axis representing frequency and the x-axis representing time. And it is widely recognized that LSTM does a very good job of drawing statistical inferences from the said time series. A transformer can also fit the job description very well, but it will be an overhaul of computational resources, and simply unnecessary. Transformers are only economically ideal when it comes to extremely long sequences. But in our case, the sequence length per time step is only forty.

Convolutional Networks (CNNs) show the most promise as with most computer vision-related tasks. The performance of CNNs in general computer vision tasks improves when the architecture is made deeper or wider. This trend also translates to the task of facial emotion recognition. The VGG-16 is a CNN with a depth of 16 layers that won ILSVRC (Imagenet), 2014. One may load the pre-trained instance of the network that has been trained on over a million photos from the ImageNet database. The pre-trained network can categorize photos into thousands of different categories. As a consequence, the network has learnt rich feature representations for a diverse set of pictures. The network's picture input size is 224 by 224.

However, experimentally we found that the Places-365 version of VGG16 does a better job of facial emotion recognition than the Imagenet variant. This can primarily be attributed to the fact that the Places365 database (which is the latest version of the Places2 Database) does a better job of training CNNs for scene recognition. That said, since we are extracting the deep scene features from the higher-level layers of VGG-16-Places-365, they augment the performance in identifying generic features for facial feature recognition useful for emotion identification.

### 3.1.1 Dataset

For this purpose, we used a combination of four datasets for emotion recognition in audio files, and one for emotion recognition from video (processed as still photo frames in practicality). All combinations of these fundamentally had the same 7 cardinal emotions: anger, neutral/contempt, disgust, fear, sadness, happiness, and surprise. The datasets used for audio were RAVDESS [25], CREMA-D [26], SAVEE [27] and TESS [28]. For video (still photographs), we used CK+ [29].

### 3.1.2 Model

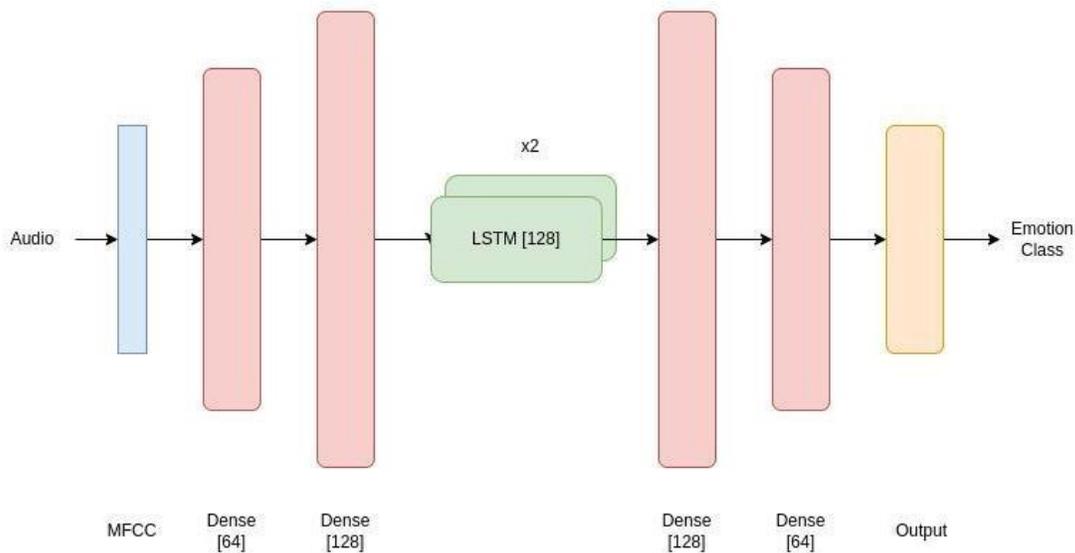

Figure 3. Model Pipeline for Emotion Detection from Audio Features

The audio input is processed and Mel frequency cepstral coefficients (MFCCs) are extracted with 40 dimensions. These are stored in a NumPy data array as input variables. The labels are stored similarly to the output variables. The model consists of eight layers. The first one is a standard input layer of dimensionality [1,40]. The next is a dense layer of 64 nodes, followed by another dense layer with 128 nodes. To increase the feature space, thus allowing for more efficient backpropagation and better learning, we tend to use dense layers that connect the input layer to our next and probably the most important segment of the model - the LSTMs. Now, as to why we use different sizes of the layers, if we directly scale up from 40 to 128 just to increase the feature space, we noticed that using two fully connected layers with 128 nodes between the input node and the LSTMs led to significant instability and a drop in the accuracy metric. Thus, we take a more measured and graded approach.

As discussed before, the inputs from the dense layer with 128 nodes are fed to two LSTM layers each with 128 nodes and sent forward to two other dense layers with 128 and 64 nodes respectively. This approach towards using two LSTMs stacked up on one another is called a hierarchical or stacked LSTM model. This allows for the hidden states from the first LSTM to propagate into the second. Stacking the LSTM layers deepens the model, more correctly defining it as a deep learning approach. The depth of neural networks is often ascribed to the approach's performance on a wide range of difficult prediction tasks. And in this case, although it makes the model more computationally expensive, it also gives us the required accuracy boost. As for the usage of the dense layers after the stacked LSTMs, it is because the output of an LSTM cannot directly be passed into a softmax activation function. They usually also output the hidden internal state *h*, equating the number of units to the dimensionality of this output. This is usually not the desired dimensionality of the required output layer, which is seven in our case. When we specify a dense layer(s) after the LSTM(s), it corresponds to

$$y(t) = W * h(t) \qquad (1)$$

where y(t) are the logits one needs to pass to a softmax layer, and W is simply the weight connection matrix of this last layer. As before, the graded approach in the reduction of node size in the dense layers allows us to stem instability in the model, and gradually bring the node size down to the required cardinality of the output layer.

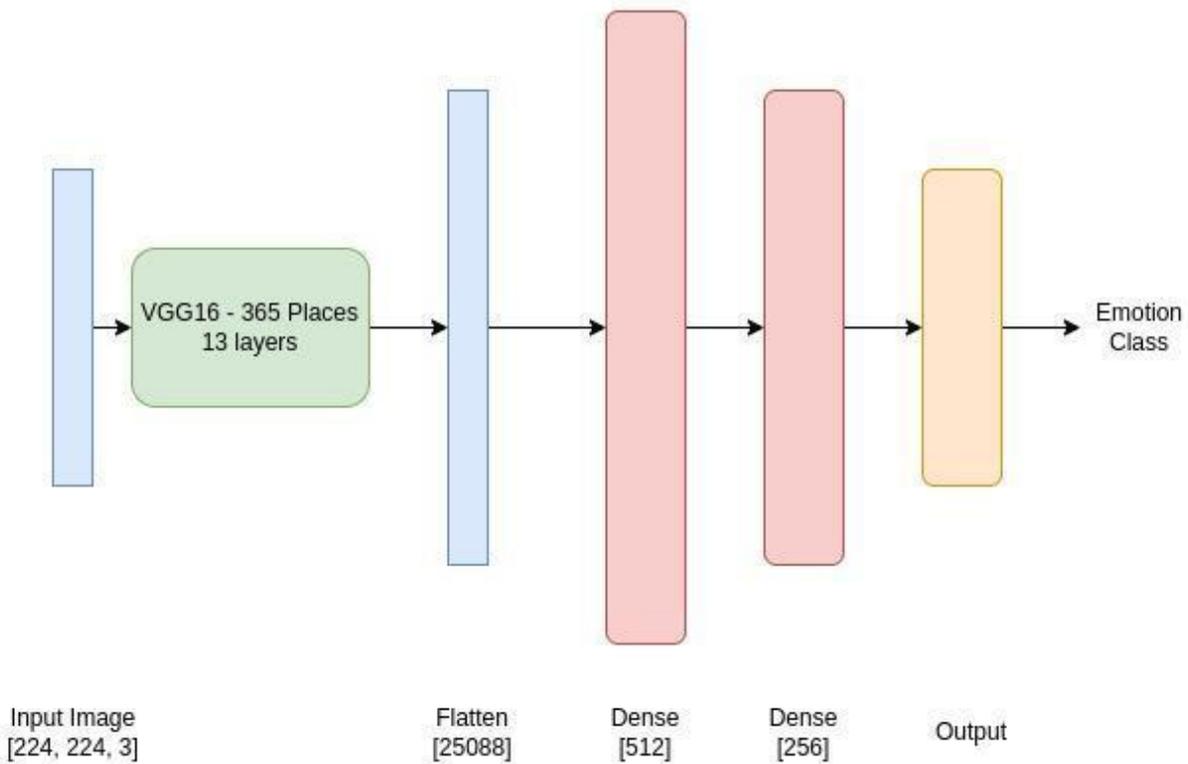

Figure 4. Model Pipeline for Emotion Detection from Video Features

The final layer is another dense layer with 7 nodes, which constitutes our output layer. For the output layer, softmax is used, and for the rest of the dense layers, ReLU is used as the activation function. The standard dropout used is 20%. This is depicted pictorially above in Figure 3.

As for the emotion detection from video, we make use of the VGG16 architecture. In our work, we used the VGG16-places365 model with no top layers (using the first 13 layers of the model). After using a Haar Cascade for facial feature detection, the images were processed as 224x224 in pixel dimensionality. The output of the VGG16 layers was flattened and then fed to a dense layer with 512 nodes, ReLU activation, and a drop out of 20%. Then it was again fed to a dense layer of ReLU activation, but consisting of 256 nodes. This is then finally passed to our output layer with 7 nodes and softmax activation, Figure 4. The dense layers before the final output layer function in a similar way to the audio emotion detection model described above. The relatively gradual decrease in layer cardinality of the nodes helps in reducing instability. These also aid the VGG-16 flattened output by enhancing the classification power of the model as a whole.

To address the situation where we may get output registered from both the emotional detection models, we use a ranking system to choose the emotion with the highest priority as the final registered output, as seen in Table 1.

Table 1: Priority rank for emotions

| Rank | Emotion |
| --- | --- |
| 1 | anger |
| 2 | disgust |
| 3 | fear |
| 4 | sadness |
| 5 | surprises |
| 6 | happiness |
| 7 | neutral/contempt |

## 3.2 Time Difference of Arrival (TDOA) Detection

Sound localization is imperative in this work as to be able to close the gap between a real dog and a robotic construct, the robot must be able to intuitively be able to respond to the vocal prompts by turning around to face the direction of the speaker. Most of the work done in this regard works on the principle that sound localization can be achieved by using an array of microphones that pick up the sound waves that bounce back off the walls and objects. By capturing the sound using different microphones, the Time Difference of Arrival (TDoA) [30], [31] can be calculated to find the direction and distance from the sound source. TDoA itself is quite a simple concept which uses the time difference at which the different microphones in the microphone array pick up the same sound. Since the distance and angles between the microphones themselves are known, it is possible to calculate the distance and the direction of the point of the origination of the sound, Figure 5.

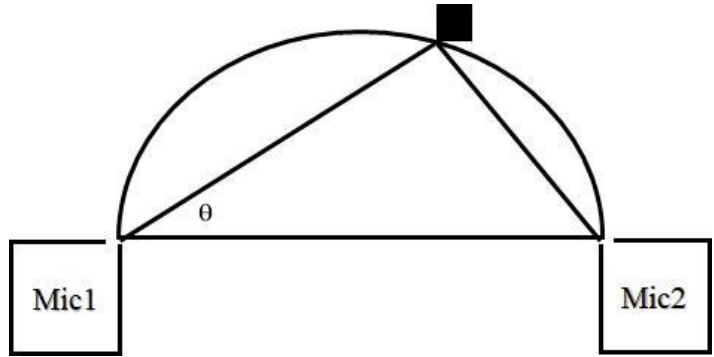

Figure 5. TDOA for two microphones

The calculation of TDoA can be done with only two microphones, however more are necessary to determine whether the sound source is originating from in front of the array or from the back. To this end, we use four microphones arranged in a cross-like manner, Figure 6, which allows the simulated robot to successfully identify the distance and direction of the sound originating anywhere. TDoA is implemented by starting a timer as soon as any one of the microphones picks up a sound signal and the timer stops when the microphone directly opposite to the first microphone picks up the same sound. The resulting time is the TDoA between the pair of microphones, this can then be used in the following formulae to calculate the angle from the first microphone from which the sound is originating.

$$\theta = arccos(\frac{\Delta t * v}{d}) \qquad (2),$$

where $\theta$ is the angle of elevation from mic 1 to the source of sound (considering, mic 1 and mic 2 lie on the x-axis), $\Delta t$ is the difference of arrival times (for sound) between the two microphones, $v$ is the velocity of sound, and d is the shortest distance between them.

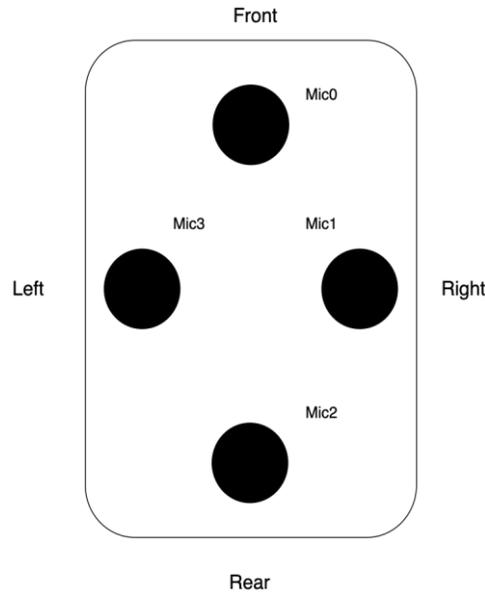

Figure 6. Placement of microphones (Seen from top view)

## 3.3 Gait Generation

In this work, we use a two-joint leg structure along with the implementation of Proximal Policy Optimization (PPO)[32] for the actual gait generation. As seen from the works done by Hengst, Ibbotson, Pham, et al.[16] and Kim & Uther [17], one can calculate and decide on a locus for the path of the robots' legs in the simulation, however, such practice is quite impractical. This is because a locus will not be able to adjust for the variance in terrain surfaces smoothly. For this reason, this work uses the PPO algorithm, a popular Reinforcement Learning method that trains the model by learning the experience from small batches of the training simulation. The learnt experience is then used to update the decision-making policy. Once the model has gone over the entire train set, the learnt mini-batch experiences are discarded and a new batch is trained upon using the updated policy. This "on-policy learning" ensures that information from bad training batches does not propagate Forward too much, making an awkward or unusable gait for the quadruped. While this does mean that there is a lot less variance in the training, the final result will be a much smoother training process and also ensure that the model does not develop any habits in the gait that cause senseless actions.

In model-based RL a model of the environment is learned, which enables the agent to plan. The transition probability distribution and the reward function constitute the model. Model-free RL manages to infer without such a model. Model-based RL enables agents to weigh available actions and their implications explicitly. This allows an agent to plan and leads to sample efficiency during training. It was successfully used in AlphaZero, a program that mastered e.g. Go or Shogi. However, in many application cases, it seems to overfit to the point where models completely fail in a real environment. While being a less sample-efficient model-free RL is easier to implement and less prone to overfitting.

Model-free RL is further divided into the families of Policy Optimization and Q-Learning. With Q-Learning an Optimal Q-function is estimated and optimized. While being more sample-efficient, performance stability is dependent on how well the Optimal Q-function can be estimated. Policy Optimization on the other hand optimizes the agent performance directly, resulting in more stable and reliable performance. For Policy Optimization a policy is explicitly represented and optimized to maximize return.

This paper uses the Proximal Policy Optimization (PPO) algorithm with a hybrid policy defined as a(t) - user policy, π(o) – feedback. This hybrid policy allows the model to be changed easily from being fully user-specified to completely dependent on the gait learnt whilst training. For example, for a completely user-defined model, setting the feedback component's upper and lower bounds to 0, we ensure that no information that is learnt while training is propagated to the next train batch. An OpenAI gym environment, is used to learn how to gracefully assume a pose avoiding too fast transactions. OpenAIs Gym allows us to easily create and use a physics environment based on the simulation. While Gym does have many pre-existing environments for many different use-cases, we needed to modify the environment till we stripped everything but the physics. We then make a new actor to represent the dog, which our PPO model then controls. It uses a one-dimensional action space with a feedback component π(o) with bounds [-0.1, 0.1]. The feedback is applied to a sigmoid function to orchestrate the movement. When the --playground flag is used, it's possible to use the pyBullet UI to manually set a specific pose altering the robot base position (x,y,z) and orientation (roll, pitch, jaw).

### 3.4 Audio-Visual Feedback

The feedback is provided through rather elementary software engineering. An LCD and a speaker output feedback mapped to each emotion. Although rather trivial, we believed it to be a necessary touch to offer a more natural final product.

## 4. Experimental Setup

As the primary goal of this work is to create a simulation of a dog-like quadruped robot, the standard environment for gait simulation had to allow for inputs that are usually not needed. For the emotion recognition, we used a fairly simple set-up - a webcam for the video emotion recognition and a microphone for the audio emotion recognition modules. These two inputs were passed through to a modified gym environment to allow for the simulation to run inside a single system. The audio passthrough also lets us implement TDoA inside the simulation, and we implemented a point-and-click method of introducing the sound source in the environment to test the same. The terrain is simulated in the gym environment to be random, to allow for accurate testing of the robot. The TDoA also allows us to orient the robot in terms of autonomous gait generation through the multi-terrain system in the simulation. However, we did not use a particular metric to judge the quality of locomotion in the random terrain environment. It was a question of whether can or cannot. As for the audio and visual feedback, these were obtained by outputting a .wav track recording of dog sounds corresponding to the emotion inferred and a Tkinter GUI screen to show a loop of translated LCD controller code to recreate the images we made to show the robot's emotion (in response to the emotion inferred) in an LCD screen set on what analogically would be its face.

A four-microphone system is used to get spatial audio. Alongside the audio emotion recognition module, the inputs are fed parallelly to the TDOA module to get an estimate of the direction of the source of the sound. An RGB camera is used to take video stills, and Haar Cascade is used to extract facial features. These are then passed on to the emotion analysis module to get the video and audio emotion inference. As discussed prior, we use a hardcoded table assigning each of the seven cardinal emotions to a priority-based ranking system, in terms of how important it is to react to the emotion. Now the emotional inferences are processed and the emotion with a higher priority is acted upon (between the two most probable emotions inferred from the pair of emotion recognition modules). For non-urgent emotions, the squat position is triggered and further feedback is provided through the LCD and speaker. For a more urgent emotion inference in the simulation like anger or sadness, the robot locomotes towards the source of the sound while providing feedback through the LCD and speaker ensemble.

## 5. Results and Analysis

The emotion detection module for audio input achieved an overall test accuracy of 63.47%, Figure 7, over the compiled dataset including RAVDESS, CREMA-D, SAVEE, and TESS. The module for the emotion detection from video input did surprisingly well with an accuracy of 99.66%, Figure 8, in the CK+ dataset, beating the previous 3rd highest (in 7 emotions accuracy), and lagging behind the 2nd highest accuracy score just by roughly .04% as per the benchmark in paperswithcode.com [32], not considering approximation. We would like to make a note here, that although the FN2EN [33] tops the dataset leaderboard in paperswithcode.com, this is attributed to their highest accuracy in the 8 emotion classes category with an accuracy of 96.8%, while it has an accuracy of 98.6% in the 6 emotions category. They do not have available test scores in the 7 emotions category.

Because our fundamental end-goal is to perform the task of classification, Categorical Cross-Entropy works best for our loss function in both types of emotion recognition, viz., from facial features and speech.

Figure 7 shows that the model converges around the 48th epoch. There it stabilizes around the 60% mark of validation accuracy. But the model tends to overfit a little, shooting towards the 68% mark (in the training accuracy metric). The same is reflected in the loss function graph. The fact that the validation loss and accuracy start decoupling from the train loss and accuracy suggests that the bias-variance tradeoff starts getting skewed. After the 48th epoch, the model tends to get biassed towards the training dataset. With that said, however, experimentally the results showcased in, Figure 7, were the best results that we obtained from several different variations of the model, Figure 3, not considering the training epoch length.

Figure 8 shows spikes forming between the 20th and the 27th epochs. These anomalies are an inevitable side effect of Adam's Mini-Batch Gradient Descent (we use a batch size of 32). Some mini-batches have accidental "unlucky" tuples for the optimization, causing these and affecting the cost function and the accuracy metric. When we implemented Stochastic Gradient Descent (the same as when the batch size is one), we noticed that the cost function has even more anomalies. This does not occur, however, if we use (Full) Batch Gradient Descent. This is because it uses the training dataset entirely during each optimization epoch. However, Adam optimization does do the job better than most when the end justifies the means.

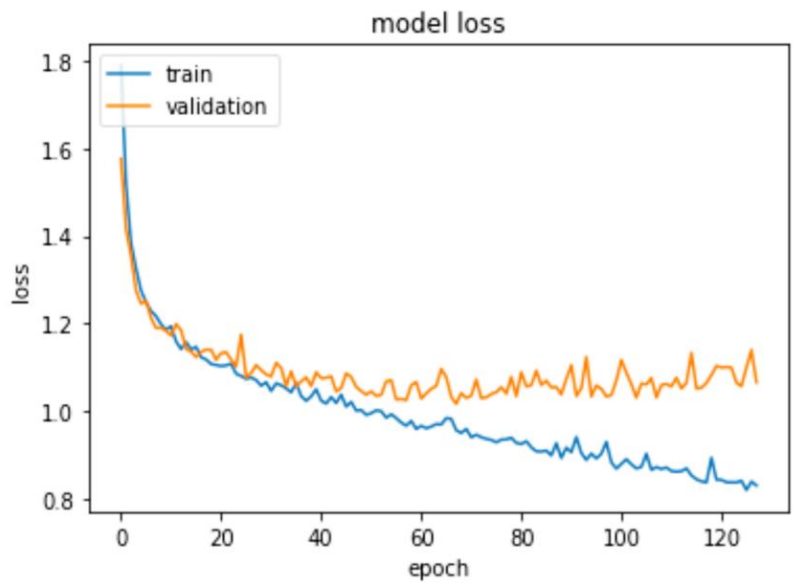

(a) Loss

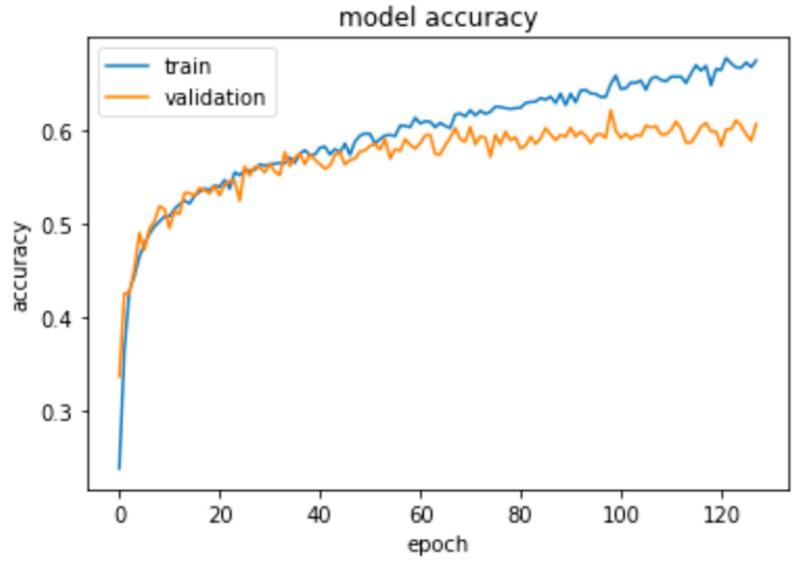

(b) Accuracy

Figure 7. Model loss and accuracy for emotion recognition from audio features

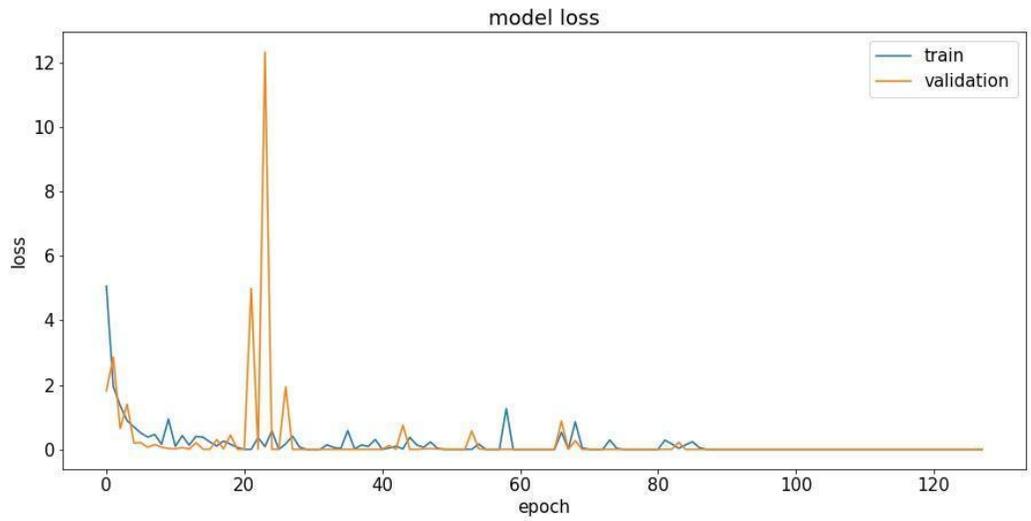

(a) Loss

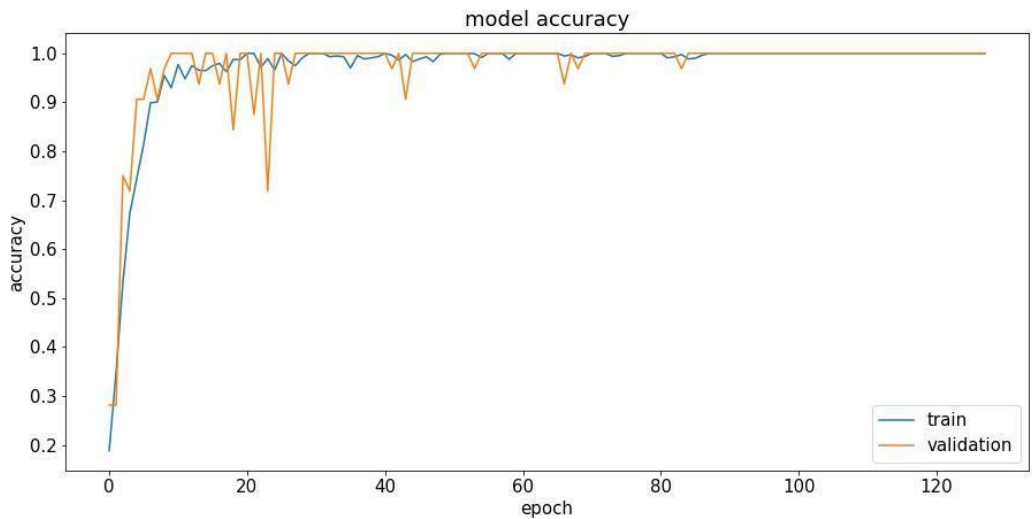

(b) Accuracy

Figure 8. Model loss and accuracy for emotion recognition from video features

The simulation for gait generation worked as expected, Figure 9, achieving locomotion in all the simulated terrains. This gym environment is used to learn how to gracefully start the different actions and then stop them after reaching the target position. Walking uses a two-dimensional action space with a feedback component π(o) with bounds [-0.4, 0.4], while for galloping we use a two-dimensional action space with a feedback component π(o) with bounds [-0.3, 0.3]. For both walking and galloping a correct start contributes to void the drift effect generated by the gait in the resulting learned policy. For standing up the action space is equal to 1 with a feedback component π(o) with bounds [-0.1, 0.1] used to optimize the signal timing. The signal function applies a 'brake', forcing the robot to assume a halfway position before completing the movement.

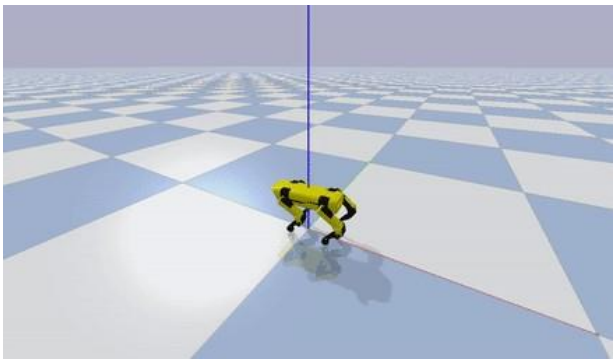

(a) Flat Terrain

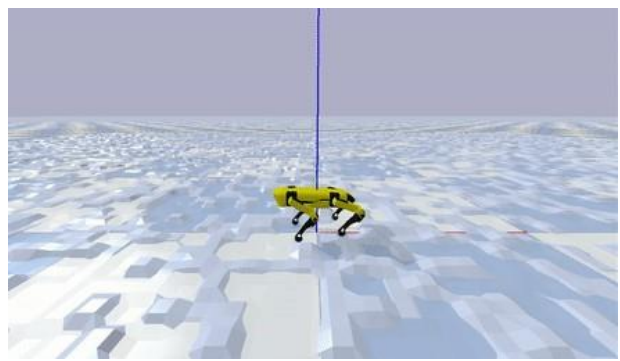

(b) Uneven Terrain

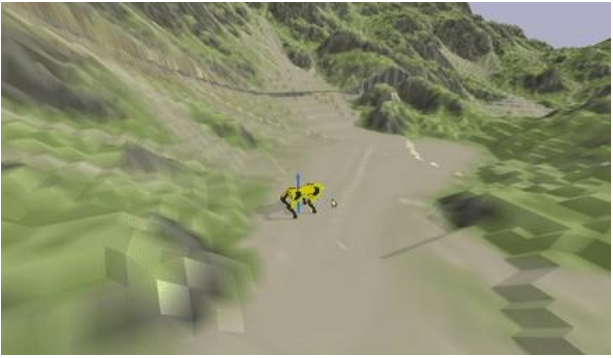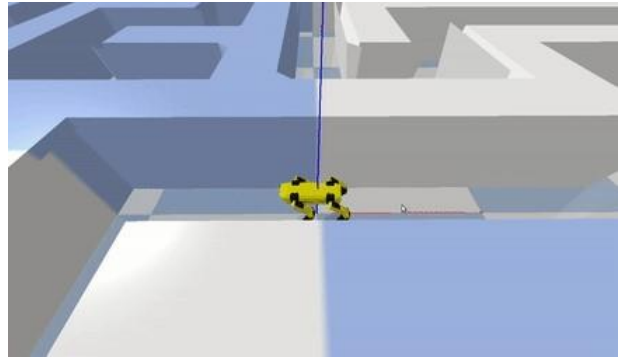

| (c) Hilly Terrain | (d) Maze |

Figure 9. Quadruped Simulation

## 6. Conclusions and Future Scope

In this section, we discuss more practical features that can be added to this quadruped if one is possessed the requisite resources and time. The primary point to focus on should be to avoid what we like to call "trip-fall-die", to explain which, we make clear that presently, we do not have any dynamic obstacle avoidance implemented in the gait generation model. In actual fabrication, one must also consider adding a gyroscopic stabilization module into the robot. To talk about the more human-centric aspects, one might consider gesture recognition using skeleton structure analysis of humans around the bot and provide necessary feedback. The requirement of responding to a name (a name assigned to the robot) and following the source of sound to the caller may also be implemented. If the video emotion inference produces an urgent emotion, it also needs to go towards the detected human. In case of medical or other emergencies, simple modules can be integrated to call emergency services as necessary. Online training can be introduced to train for patterns of behaviours, or perhaps even to go as simplistic and learn tricks that a dog could perform. A support dog module can also be included to comfort the owner in times of emotional need. Another interesting feature that we would love to see would be a sentinel mode, where the quadruped walks the perimeter of an assigned area, barks at faces unknown to its memory, and maybe even gives warning before stunning any potential criminal attempting a break into the said premises.

As discussed above, this paper aims to establish the framework for simulating a quadruped robot that is emotionally intelligent and can primarily respond to audio-visual stimuli using motor or audio response. With that said, this work was mainly construed as a proof of concept to showcase the fact that with existing work in literature we will be able to put together a system that can mimic a real dog. In our simulation, we show that the use of age-old techniques such as the MFCC algorithm can be brought back to life using newer architectures in an ensemble to complement it. This audio emotion detection performs relatively well, approximately 63.47% while needing minimal computational resources. However, it is important to note that this does not perform at par with the current works in the related datasets, viz, the ERANNs on the RAVDESS dataset with a top accuracy of 73.4% [34], or the Zeta Policy training on the SAVEE dataset, with an accuracy of

68.90 ± 0.61%. On the other hand, the video emotion detection system is a novel architecture that was created for this work and produces results that are almost at par with the state of the art. The accuracy attained (99.66%) is only topped by models FAN [35] and Vit + SE [36] with respective accuracies of 99.7% and 99.8%. The other main component is locomotion (automated gait generation) which is simulated in this work by using the PPO algorithm and the TDoA algorithm. The PPO algorithm is especially quick in learning due to its "on-policy" learning methodology allowing for the simulated dog to exhibit a very smooth gait through the various cadences and changes. This then allows for the simulated quadruped robot to be reactive to the simulated stimuli, thus allowing us to say that it performs as expected and meets the goal of this paper.

**Statements & Declarations:**

**Funding:** The work carried out in this research has not received any funding.
**Competing Interests:** The authors declare that they have no known competing financial interests or personal relationships that could have appeared to influence the work reported in this paper.
**Author Contributions:** All authors have contributed to the research study conception and design. Abhirup has contributed to the concept ideation. Jatin has contributed to the concept implementation. The whole work was supervised by Sibi Chakravarthy and Aswani Kumar. The first draft was prepared by Anitha has compared the analysis. Firuz provided crucial feedback and helped to interpret the results. Annapurna has provided the inputs for implementation, verified the numerical results and coordinated.
**Ethics Approval:** This work do not deal with any research involving either human or animal subjects. Hence this approval is not relevant to this article.
**Consent to participate:** This work do not deal with any research involving either human or animal subjects. Hence this approval is not relevant to this article.
**Consent to publish:** This work do not deal with any individual data and hence this consent is not relevant for this article.
**Acknowledgements:** NIL
**Code / Data Availability Statements:** Code / Data sharing not applicable to this article as no datasets were generated or analyzed during the current study.